%% file: root.tex
\documentclass[letterpaper, 10pt, conference]{ieeeconf}
\IEEEoverridecommandlockouts
\usepackage{cite}
\usepackage{amsmath,amssymb,amsfonts}
\newcounter{remark}
\newenvironment{remark}{%
  \refstepcounter{remark}%
  \par\smallskip\noindent\textbf{Remark~\theremark.}\ }{%
  \par\smallskip}
\usepackage{algorithm}
\usepackage{algpseudocode}
\usepackage{graphicx}
\usepackage{textcomp}
\usepackage{xcolor}
\usepackage[final]{changes}
\definecolor{RevisionAdd}{RGB}{0,110,70}
\definecolor{RevisionDelete}{RGB}{170,45,45}
\setaddedmarkup{\textcolor{RevisionAdd}{\uline{#1}}}
\setdeletedmarkup{\textcolor{RevisionDelete}{\sout{#1}}}
\usepackage[caption=false,font=footnotesize]{subfig}
\usepackage{booktabs}
\usepackage{multirow}
\usepackage{stfloats}
\usepackage{url}
\def\BibTeX{{\rm B\kern-.05em{\sc i\kern-.025em b}\kern-.08em
    T\kern-.1667em\lower.7ex\hbox{E}\kern-.125emX}}
\begin{document}

\title{MetaTune: Adjoint-based Meta-tuning via Robotic Differentiable Dynamics}

\author{%
Xiexin Peng$^{1}$, Bingheng Wang$^{2}$, Tao Zhang$^{1,*}$, Que Dong$^{3}$, and Ying Zheng$^{1,*}$%
\thanks{$^{1}$\,Xiexin Peng, Tao Zhang, and Ying Zheng are with the School of Artificial Intelligence and Automation, Huazhong University of Science and Technology, Wuhan 430074, China.}%
\thanks{$^{2}$\,Bingheng Wang is with the Department of Electrical and Computer Engineering, National University of Singapore, Singapore 117583, Singapore.}%
\thanks{$^{3}$\,Que Dong is with Wuhan Zhenyou Technology Co., Ltd., Wuhan 430205, China.}%
%\thanks{$^{4}$\,Ying Zheng is with Shenzhen Huazhong University of Science and Technology Research Institute, Shenzhen 518057, China.}%
\thanks{$^{*}$\,Corresponding author: Tao Zhang(taozhang\_mse@hust.edu.cn), Ying Zheng (zyhidy@mail.hust.edu.cn).}%
\thanks{This work was supported in part by the National Natural Science Foundation of China under Grant 62503188, in part by the Hubei Province International Science and Technology Cooperation Project under Grant 2024EHA033, and in part by the Guangdong Basic and Applied Basic Research Foundation under Grant 2025A1515010134.}%
}

\maketitle
\thispagestyle{empty}
\pagestyle{empty}

\begin{abstract}
Disturbance observer-based control has shown promise in robustifying robotic systems against uncertainties. 
However, tuning such systems remains challenging due to the strong coupling between controller gains and observer parameters. 
In this work, we propose MetaTune, a unified framework for joint auto-tuning of feedback controllers and disturbance observers through differentiable closed-loop meta-learning. MetaTune integrates a portable neural policy with physics-informed gradients derived from differentiable system dynamics, enabling adaptive gains across tasks and operating conditions. 
We develop an adjoint method that efficiently computes the meta-gradients with respect to adaptive gains backward in time to directly minimize the cost-to-go. 
Compared to existing forward methods, our approach reduces the computational complexity to be linear in the data horizon.
On quadrotor control tasks, MetaTune achieves competitive or improved tracking performance while reducing gradient computation time by more than 50\%.
In PX4-Gazebo hardware-in-the-loop simulation, the learned policy transfers zero-shot and reduces tracking RMSE by about 15--20\% in aggressive flight and up to 40\% under strong disturbances.
\end{abstract}


\section{Introduction}

As robotic systems extend beyond structured environments into complex and uncertain real-world settings, control design faces fundamentally new challenges. Applications such as offshore UAV inspection and rough-terrain locomotion demand robustness against stochastic disturbances and modeling uncertainties.
To meet these demands, integrating feedback controllers with disturbance observers has become a widely adopted strategy in robust control design.
This paradigm builds upon established estimation frameworks, extending from the classical Kalman filter\cite{kalman1960new} and Luenberger observers\cite{luenberger2007observing} to robust architectures including disturbance observers (DOBs), extended state observers (ESOs), and sliding-mode observers (SMOs)\cite{sariyildiz2019disturbance, han2009pid, gao2003scaling, spurgeon2008sliding}.

\begin{figure}[t]
    \centering
    \includegraphics[width=\columnwidth]{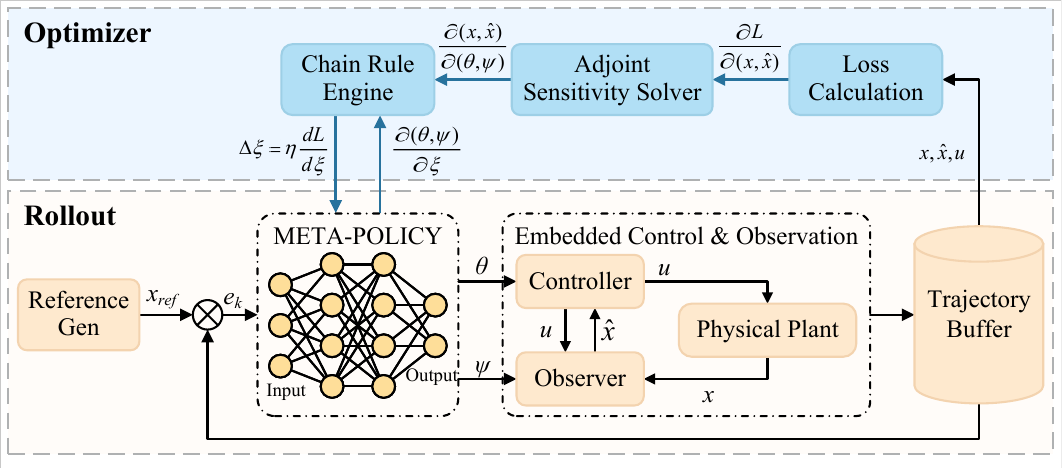}
    \caption{
    Adjoint-based meta-learning framework for end-to-end tuning of controller--observer parameters.
    Gradients of a task-level loss are propagated backward through the closed-loop system dynamics via adjoint sensitivity analysis and subsequently transformed into meta-policy gradients using the chain rule, enabling automatic gain adaptation.
    }
    \label{fig:overview}
\end{figure}

Observer-based control architectures critically depend on coordination between control and estimation parameters for nonlinear systems. Control strategies based on static assumptions or fixed parameters are inherently limited when confronted with time-varying, nonlinear disturbances that are challenging to model explicitly\cite{zhang2023unknown}. 
To address this issue, variable-gain strategies such as gain scheduling, LQR-based tuning \cite{leith2000survey}, and fuzzy control \cite{mamdani1974application} have been explored. However, these approaches typically rely on offline-designed rules and lack true online adaptability under uncertainty.
Adaptive control offers a principled alternative by adjusting controller and observer gains based on real-time system behavior. Classical MRAC ensures reference tracking with Lyapunov stability guarantees \cite{slotine2002adaptive}, and similar ideas have been extended to observer-based architectures for disturbance compensation \cite{marine2001robust}.

Despite extending the applicability of fixed-gain structures, these methods fundamentally depend on predefined operating conditions or expert-designed adaptive laws, limiting their ability to generalize under unknown disturbances and evolving environments. This limitation has motivated learning-based approaches that leverage data-driven representations to enhance robustness beyond model-based adaptive control assumptions.

Recent learning-based approaches for robust control can be broadly categorized into two distinct groups based on their intervention mechanisms. The first category focuses on disturbance-aware learning. Representative works explicitly estimate external disturbances or unmodeled dynamics for compensation in the control loop \cite{joshi2019deep, he2025neural, giovanni2023hdvio, o2022neural, pan2026learning}, while other approaches implicitly account for disturbance effects by updating control policies directly based on closed-loop performance metrics \cite{richards2023control}.
The second category concentrates on intelligent parameter adaptation, where learning algorithms adjust controller or observer gains online to improve closed-loop performance. Reinforcement learning methods formulate gain tuning as a reward maximization problem \cite{chen2020q, bujgoi2025tuning}. Meta-learning approaches exploit shared structure across tasks to enable rapid adaptation and improved generalization \cite{wei2025meta}. 
More recently, the integration of differentiable physics has formalized parameter tuning as a gradient-based optimization problem. By modeling system dynamics as differentiable computational graphs, frameworks such as DiffTune \cite{cheng2024difftune} and Neural MHE \cite{wang2023neural} propagate analytical gradients of task-level losses directly through the control loop, which facilitates end-to-end optimization of controller or estimator parameters using trajectory data, thereby effectively mitigating the sample inefficiencies associated with black-box search methods.

However, existing methodologies face a fundamental limitation in terms of the interplay between control and estimation during the learning and adaptation process. First, disturbance-aware approaches primarily emphasize the explicit compensation of external forces, often overlooking the optimization of the feedback controller itself. Even when augmented with robust mechanisms such as $\mathcal{L}_1$ adaptive control\cite{hovakimyan2010L1}, these methods remain constrained by manually designed reference models or fixed filter bandwidths, which restrict their adaptability to dynamic variations. Second, while parameter adaptation strategies efficiently address gain tuning, they typically treat control and estimation as isolated processes. This decoupled perspective fails to acknowledge the intrinsic coupling in disturbance-observer-based architectures. Consequently, such isolated tuning schemes fail to balance the trade-off between disturbance rejection and noise sensitivity, potentially leading to sub-optimal or even unstable closed-loop behavior.

In this work, we propose MetaTune, a unified framework that achieves joint tuning of feedback control and disturbance estimation by formulating the tuning problem as a differentiable optimization over closed-loop executions. As shown in Fig.~\ref{fig:overview}, MetaTune integrates a portable neural network with differentiable control and observation modules to generate adaptive gains across tasks. To enable efficient and scalable learning, we compute meta-gradients by solving adjoint dynamics backward in time, which yields vector-valued gradient trajectories.
Compared with DiffTune \cite{cheng2024difftune}, which relies on forward sensitivity propagation, the proposed adjoint-based approach computes trajectory-level gradients in a single backward sweep. As a result, the gradient computation complexity scales linearly with the horizon length, rather than quadratically, and remains independent of the parameter dimension.
This property is particularly advantageous when observer dynamics are incorporated into the tuning process, where the augmented state and gain dimensions substantially amplify the computational cost of forward sensitivity methods.

The main contributions of this paper are summarized as follows:
\begin{itemize}
    \item \textbf{Unified Joint Tuning.}
    A unified architecture is proposed that embeds differentiable system dynamics into the learning policy, enabling end-to-end joint optimization of control and estimation for robust adaptation.
        
    \item \textbf{Scalable Adjoint-based Method.} \replaced{An adjoint-based meta-gradient method is derived for the coupled closed-loop system, enabling efficient gradient computation with cost independent of the number of gain parameters and supporting high-dimensional gain parameterizations.}{ 
    An adjoint-based meta-gradient method for joint tuning is derived, achieving parameter-dimension--independent computational complexity and scalability to high-dimensional gain parameterizations.}

    \item \replaced[comment={R6404-C1}]{\textbf{High-Fidelity Quadrotor Validation.}
    MetaTune is validated in high-fidelity quadrotor simulation with a lightweight MLP-based gain-adaptation policy, demonstrating robust transfer across unseen trajectories and disturbances.}{\textbf{Robust Generalization.} 
    MetaTune is validated on robust quadrotor flight control in high-fidelity simulation. 
    A portable policy parameterized by a single-hidden-layer multilayer perceptron (MLP) is employed for gain adaptation. 
    The proposed framework demonstrates robust generalization across unseen trajectories and disturbances.}
    
\end{itemize}

The rest of this paper is organized as follows. 
Section II introduces the problem setting.
Section III presents the proposed adjoint-based meta-gradient computation framework.
Section IV presents experimental evaluation of the proposed method on a quadrotor platform. 
Section V discusses the implications and limitations of the approach and concludes the paper.
The source code is available at \url{https://github.com/p-xiexin/MetaTune}.

\section{Problem Statement}
\label{sec:problem}

Consider a discrete-time nonlinear dynamical system operating in a complex environment subject to generalized disturbances. The evolution of the system state is governed by the open-loop dynamics:
\begin{equation}
    x_{k+1} = f(x_k, u_k) + d_k,
    \label{eq:dynamics}
\end{equation}
where $x \in \mathbb{R}^n$ represents the system state, $u \in \mathbb{R}^m$ is the control input, $d \in \mathbb{R}^{n}$ denotes the generalized disturbance vector, and $k$ signifies the time step. The function $f: \mathbb{R}^n \times \mathbb{R}^m \to \mathbb{R}^n$ represents the nominal model of the system.

\begin{remark}
The term $d_k$ represents a generalized abstraction of system uncertainties. It aggregates the effects of external perturbations and modeling inaccuracies resulting from discrepancies between the actual system dynamics and the nominal model. Importantly, the disturbance sequence $\{d_k\}$ is unknown and unstructured: it is neither directly measurable nor subject to optimization. The influence of disturbances within systems must be mitigated to achieve satisfactory control performance.
\end{remark}

To offset the influences of disturbances within systems, observer-based control methods are commonly employed.
The system state is partially accessible through the measurement equation $y_k = g(x_k)$, where $y_k \in \mathbb{R}^p$ denotes the measured output corrupted by sensor noise. To simultaneously reconstruct the system state and compensate for the lumped disturbance, the disturbance observer is constructed with an augmented state representation as follows:
% \begin{equation}
%     \hat{z}_{k+1} = o(\hat{z}_k, u_k, y_{k+1}; \psi_k),
%     \label{eq:observer}
% \end{equation}
\begin{equation}
    \hat{x}_{k+1} = o(\hat{x}_{k}, u_k, y_{k+1}; \psi_k),
    \label{eq:observer}
\end{equation}
where we denote the augmented observer state as 
$\hat{x}_k = [\hat{x}_k^{\,\mathrm{sys}},\ \hat{d}_k]^\top$, which comprises both the estimated system state and disturbance. Moreover, $\psi_k \in \mathbb{R}^{n_\psi}$ represents the time-varying observer parameters.

In observer-based control methods, the control input is calculated using the estimated state through a parameterized feedback law tracking a reference trajectory $x_k^d$:
\begin{equation}
    u_k = h(\hat{x}_{k}, x_k^d; \theta_k),
    \label{eq:controller}
\end{equation}
where $\theta_k \in \mathbb{R}^{n_\theta}$ denotes the time-varying controller gains.
We assume that $f, g, o$, and $h$ are strictly differentiable with respect to their state and parameter arguments.

With the system architecture in place, we reformulate the parameter adaptation process as a differentiable trajectory optimization problem. Instead of treating the parameters as static constants determined through offline tuning or isolated stability analyses, the sequences 
$\{\theta_k\}$ and $\{\psi_k\}$ are regarded as decision variables that influence the closed-loop evolution over a finite horizon. The goal is to minimize a cumulative cost $J$ subject to the explicit constraints induced by the coupled feedback and estimation dynamics.

Mathematically, the joint tuning task can be cast into the following constrained nonlinear optimization form:
\begin{equation}
\begin{aligned}
& \underset{\{\theta_k, \psi_k\}_{k=0}^{N-1}}{\text{minimize}}
& & J = \sum_{k=0}^{N-1} \ell(\hat{x}_{k}, x_k, u_k) \\
& \text{subject to}
& & x_{k+1} = f(x_k, u_k) + d_k, \\
& & & y_{k+1} = g(x_{k+1}), \\
& & & \hat{x}_{k+1} = o(\hat{x}_{k}, u_k, y_{k+1}; \psi_k), \\
& & & u_k = h(\hat{x}_{k}, x_k^d; \theta_k), \\
& & & k \in \{0, 1, \dots, N-1\}.
\end{aligned}
\tag{P}
\label{prob:optimization}
\end{equation}

Additional constraints on the parameter sequences, such as bounds or regularization terms, can be incorporated into this formulation without altering the proposed framework.

In linear time-invariant systems, estimator and controller designs can be decoupled without loss of optimality. 
In contrast, Problem~(P) considers a setting in which this separation principle does not apply, as the controller and observer are jointly tuned as components of a coupled closed-loop dynamical system. 
In this formulation, the controller operates on the observer output rather than the true state, so that the effective plant seen by the feedback law implicitly incorporates the estimator dynamics. 
As a result, control and estimation cannot be treated as independent design modules, and solving~\eqref{prob:optimization} requires identifying a coordinated solution in the joint parameter space that optimizes overall closed-loop performance.

\section{Adjoint-based Meta-Learning Framework}
\label{sec:method}

This section presents an adjoint-based meta-learning framework that exploits differentiable closed-loop dynamics to compute gradients via adjoint equations, jointly optimizing controllers and observers through a unified policy that produces task-dependent gains, enabling physics-informed end-to-end training without black-box methods.

% \subsection{Control-oriented Meta Learning}
% \label{subsec:meta_learning}
\deleted{We adopt a control-oriented meta-learning perspective, treating time-varying controller and observer gains as policy outputs. }
\deleted{Each task $\mathcal{T}$ is sampled from a distribution $p(\mathcal{T})$ characterizing environment and system variations. Distinct tasks induce unique closed-loop behaviors, necessitating task-specific parameter trajectories.}
\deleted{Policy search methods address the limitations of traditional adaptive control by formulating adaptation as gradient-based optimization over parameterized policies, rather than heuristic single-task tuning.}

\deleted{A policy $\pi$ is evaluated by its expected closed-loop performance over the task distribution,}

\deleted{where \( \mathcal{\pi} \) maps task features \( \mathcal{T}^i \) to adaptive parameters, enabling task-specific tuning.}

\subsection{Neural Parameterization of Adaptive Gains}
\label{subsec:network}

We adopt a meta-learning perspective~\cite{richards2023control} by employing a neural network as the policy representation to navigate the complexity of task-dependent parameter trajectories:
\begin{equation}
\label{eq:nn}
    \eta_k = \pi_\xi (\mathcal{T}_k),
\end{equation}
where $\xi$ denotes the neural network parameters and $\mathcal{T}_k$ represents task features influencing system behavior. 
\deleted{We assume $\mathcal{T}_k$ is measurable or provided by a higher-level planner and remains exogenous to the closed-loop dynamics.}
The latent output $\eta_k$ is mapped to bounded gains $\Theta_k$ through element-wise scaling (e.g., sigmoid scaling) to satisfy predefined safety limits.

\added{The task feature $\mathcal{T}_k$ is treated as an exogenous input to the closed-loop dynamics, either measured directly or supplied by a higher-level planner. The policy is trained to minimize the expected closed-loop cost over the task distribution:}

\begin{equation}
    \mathcal{J}(\pi) =
    \mathbb{E}_{\mathcal{T} \sim p(\mathcal{T})}
    \left[
        \sum_{k=0}^{N-1}
        \ell \left(
            \pi(\mathcal{T}, k), \mathcal{T}
        \right)
    \right].
    \label{eq:meta_objective}
\end{equation}

The joint tuning problem is then formulated as finding the optimal neural network parameters \( \xi^* \) that minimize the closed-loop loss function defined in \eqref{eq:meta_objective}:
\begin{equation}
\begin{aligned}
    \min_{\xi} \quad & \mathcal{J}\big( \mathbf{\hat{x}}(\xi), \mathbf{x}(\xi), \mathbf{u}(\xi)  \big) \\
    \text{s.t.} \quad & \text{Closed-loop dynamics driven by } \pi_\xi.
\end{aligned}
\end{equation}
Compared to fixed gain tuning, the policy flexibly adjusts controller and observer parameters for different task features $\mathcal{T}_k$, enabling adaptation within a unified framework.

To update the network parameters \( \xi \), we combine the physics-informed meta-gradients derived in Sec.~\ref{subsec:adjont_grad} with the network's local gradients. The gradient of the total loss with respect to the network weights is computed using the chain rule:
\begin{equation}
\label{eq:chain}
    \nabla_\xi \mathcal{J}(\cdot)
    =
    \frac{1}{M}
    \sum_{i=1}^{M} \sum_{k=0}^{N-1}
    \left( \frac{\partial \eta_k^i}{\partial \xi} \right)^\top
    \frac{\partial \ell^i}{\partial \eta_k^i},
\end{equation}
where $M$ denotes the number of tasks sampled from the distribution $p(\mathcal{T})$, and $N$ is the finite rollout horizon length for each task. \( \frac{\partial \ell}{\partial \eta_k} \) is the sensitivity computed via the adjoint method, and \( \frac{\partial \eta_k}{\partial \xi} \) is the Jacobian of the neural network obtained via backpropagation. The parameters \( \xi \) are iteratively updated using a gradient-based optimizer (e.g. Adam) to minimize the closed-loop objective.

\added[comment={R6385-C2/R6570-C3}]{
Since weakly identifiable gain combinations may yield similar closed-loop responses, we construct a local Gauss--Newton approximation \(H_\Theta \approx \sum_k S_k^\top W_k S_k\) from rollout sensitivities and attenuate gradient components along its small-eigenvalue directions. This suppresses gain drift along poorly excited compensation modes and prevents unreliable parameter updates from dominating the joint controller--observer tuning process.
}

\input{algorithm_metatune}

\subsection{Gradient Computation}
\label{subsec:adjont_grad}
\begin{figure}[t]
    \centering
    \includegraphics[width=1.0\linewidth]{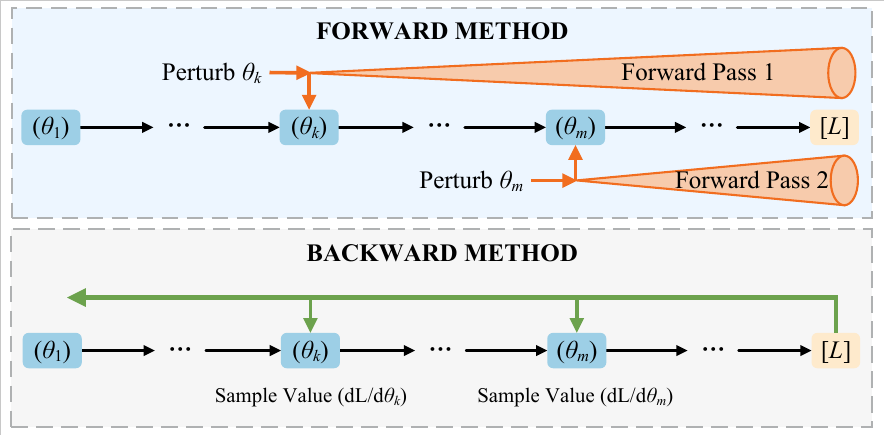}
    \caption{Computational efficiency comparison between forward and backward sensitivity analysis.
    (Top) The forward method requires a separate forward propagation rollout for each parameter perturbation to capture its effect on the final loss $L$, leading to expensive computations.
    (Bottom) The proposed backward method computes the cost-to-go function in a single backward sweep. Gradients for all parameters along the horizon are retrieved simultaneously, significantly reducing computational complexity from $O(N^2)$ to $O(N)$.}
    \label{fig:grad_compare}
\end{figure}

In the coupled closed-loop dynamics of Problem~(P), parameter variations influence the entire rollout, rendering the trajectory-level objective inherently non-local. A direct approach to compute gradients is forward sensitivity analysis. Writing the closed-loop dynamics compactly as $\mathbf{x}_{k+1} = \mathbf{F}(\mathbf{x}_k, \mathbf{\Theta}_k)$ with time-varying gains $\mathbf{\Theta}_k = [\theta_k, \psi_k]$, the state--parameter sensitivity $\mathbf{S}_k = \frac{\partial \mathbf{x}_k}{\partial \mathbf{\Theta}}$ evolves according to
\begin{equation}
\mathbf{S}_{k+1}
=
\frac{\partial \mathbf{F}_k}{\partial \mathbf{x}_k}\mathbf{S}_k
+
\frac{\partial \mathbf{F}_k}{\partial \mathbf{\Theta}_k}.
\end{equation}
Since each parameter block $\mathbf{\Theta}_k$ affects all subsequent states and the associated cost-to-go, computing trajectory gradients requires repeated sensitivity propagation, resulting in quadratic complexity with respect to the horizon length, as illustrated in Fig.~\ref{fig:grad_compare}.

To overcome these limitations, we adopt a backward sensitivity analysis based on discrete adjoint methods~\cite{akman2020efficient, andersson2019casadi}. 
The closed-loop system is represented as an augmented process encompassing both the plant and observer states. The control input $u_k$ is treated as a lifted independent variable subject to the constraint imposed by the controller, leading to the discrete augmented Lagrangian over the trajectory:
\begin{equation}
\label{eq:lagrangian}
\begin{split}
    \mathcal{L} = \sum_{k=0}^{N-1} \bigg[ 
    & \ell(x_k, \hat{x}_k, u_k) \\
    & + \lambda_{x,k+1}^\top \big(f(x_k, u_k) - x_{k+1}\big) \\
    & + \lambda_{\hat{x},k+1}^\top \big(o(\hat{x}_k, x_k, u_k; \psi) - \hat{x}_{k+1}\big) \\
    & + \lambda_{u,k}^\top \big(h(\hat{x}_k, x_k^d; \theta) - u_k\big) 
    \bigg],
\end{split}
\end{equation}
where $\lambda_{x,k}$ and $\lambda_{\hat{x},k}$ are adjoint variables associated with the plant and observer dynamics, respectively. Moreover, $\lambda_{u,k}$ is the control channel adjoint.

Taking the stationarity condition of the augmented Lagrangian with respect to the control input, $\nabla_{u_k} \mathcal{L} = 0$, yields the explicit expression for the control channel adjoint:
\begin{equation}
    \lambda_{u, k} = \frac{\partial \ell_k}{\partial u_k} 
    + \left(\frac{\partial f}{\partial u_k}\right)^\top \lambda_{x, k+1} 
    + \left(\frac{\partial o}{\partial u_k}\right)^\top \lambda_{\hat{x}, k+1}
    \label{eq:lambda_u},
\end{equation}
where the control channel adjoint $\lambda_{u,k}$ aggregates the marginal contribution of the control input through the immediate loss, the plant dynamics, and the observer update.

Substituting~\eqref{eq:lambda_u} into the stationarity conditions with respect to $\hat{x}_k$ and $x_k$ yields the coupled adjoint recursions:
\begin{align}
    \lambda_{\hat{x}, k} &= \frac{\partial \ell_k}{\partial \hat{x}_k} 
    + \left(\frac{\partial o}{\partial \hat{x}_k}\right)^\top \lambda_{\hat{x}, k+1} 
    + \left(\frac{\partial h}{\partial \hat{x}_k}\right)^\top \lambda_{u, k},
    \label{eq:lambda_hat} \\
    \lambda_{x, k} &= \frac{\partial \ell_k}{\partial x_k} 
    + \left(\frac{\partial f}{\partial x_k}\right)^\top \lambda_{x, k+1} 
    + \left(\frac{\partial o}{\partial x_k}\right)^\top \lambda_{\hat{x}, k+1},
    \label{eq:lambda_x}
\end{align}
where the adjoint variables are propagated backward in time and initialized at the terminal step. 
In the absence of an explicit terminal cost, the terminal adjoint values are set to zero.

These equations highlight the coupling: $\lambda_{\hat{x}, k}$ propagates through the observer and control policy, whereas $\lambda_{x, k}$ evolves through the plant and observer dynamics.
The backward costate recursion encodes the cumulative cost-to-go effect into vector-valued adjoint variables, yielding a compact gradient representation without explicitly storing or propagating large Jacobians.

Finally, since the tunable parameters $\Theta = \{\theta, \psi\}$ appear only in the explicit constraints defining the controller and observer, the parameter gradients are obtained by accumulating their contributions along the trajectory:

\begin{subequations}
\begin{align}
\nabla_\theta \mathcal{L}
&=
\sum_{k=0}^{N-1}
\left(\frac{\partial h}{\partial \theta}\right)^\top
\lambda_{u,k},
\label{eq:param_gradients_a} \\[2pt]
\nabla_\psi \mathcal{L}
&=
\sum_{k=0}^{N-1}
\left(\frac{\partial o}{\partial \psi}\right)^\top
\lambda_{\hat{x},k+1}.
\label{eq:param_gradients_b}
\end{align}
\label{eq:param_gradients}
\end{subequations}

Stacking \eqref{eq:param_gradients} yields the gain-level gradient $\nabla_{\Theta_k} L$. Since the gains are generated from the latent policy output $\eta_k$ through a smooth bounded reparameterization, the sensitivity required in \eqref{eq:chain} is obtained via the chain rule by propagating $\nabla_{\Theta_k} L$ through the Jacobian $\frac{\partial \Theta_k}{\partial \eta_k}$.

In summary, the adjoint-based formulation yields an efficient and scalable framework for gradient computation in closed-loop systems with coupled estimation and control. 
The resulting backward recursion naturally handles long-horizon dependencies and enables efficient parameter tuning in high-dimensional settings.

\begin{figure*}[t]
    \centering
    \subfloat[]{%
        \includegraphics[width=0.32\textwidth]{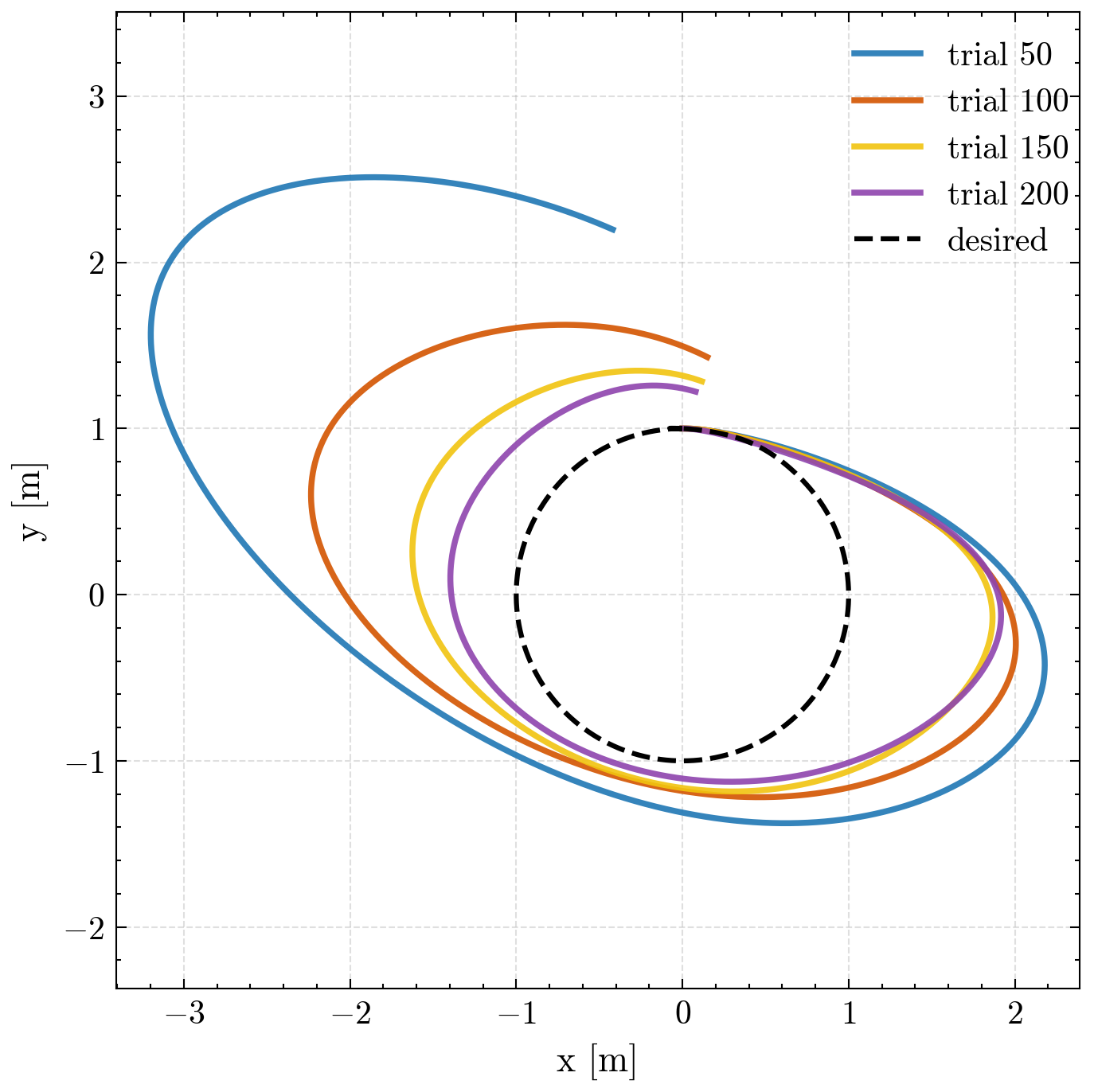}
        \label{fig:traj_a}
    }\hfill
    \subfloat[]{%
        \includegraphics[width=0.32\textwidth]{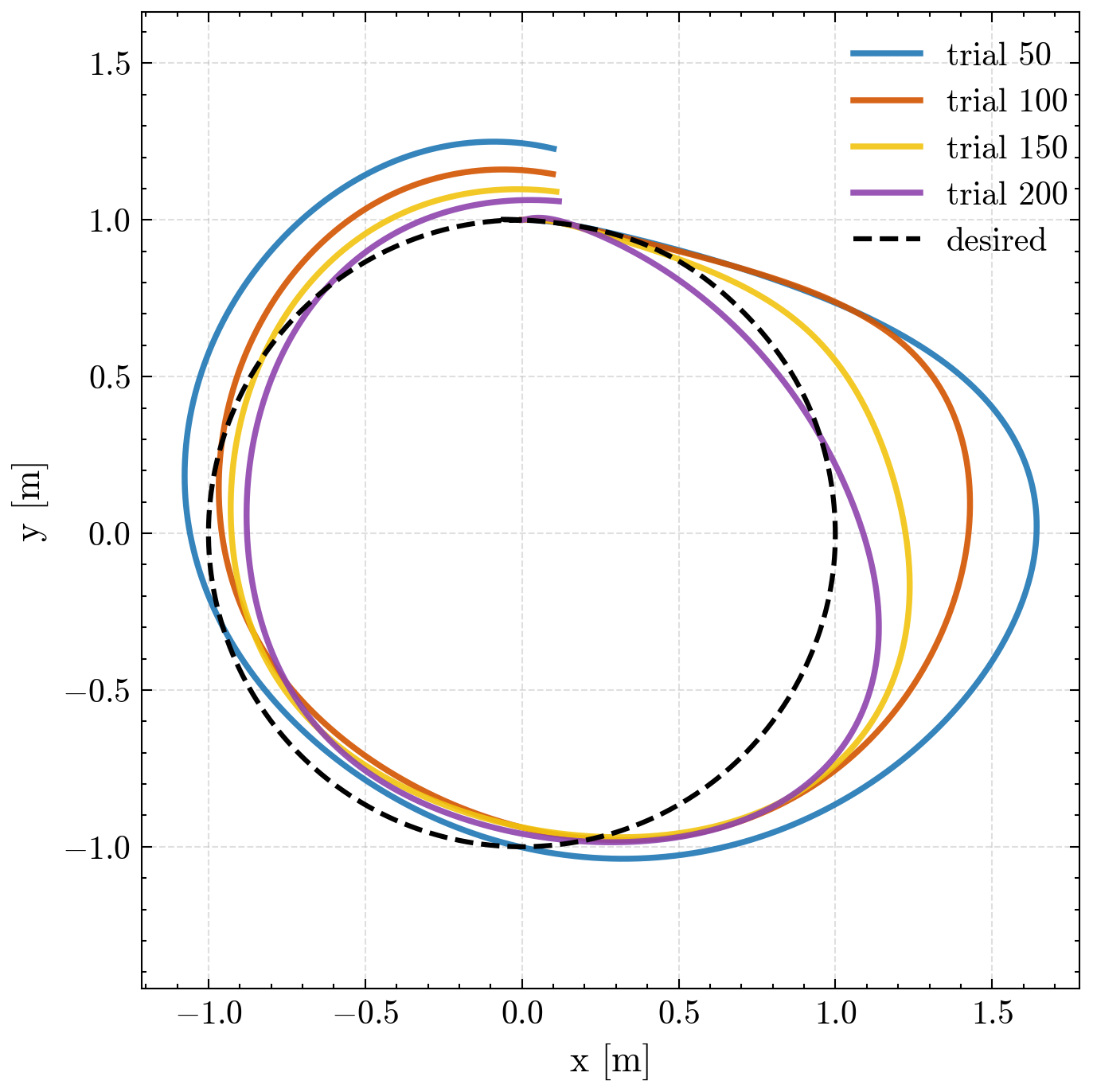}
        \label{fig:traj_b}
    }\hfill
    \subfloat[]{%
        \includegraphics[width=0.32\textwidth]{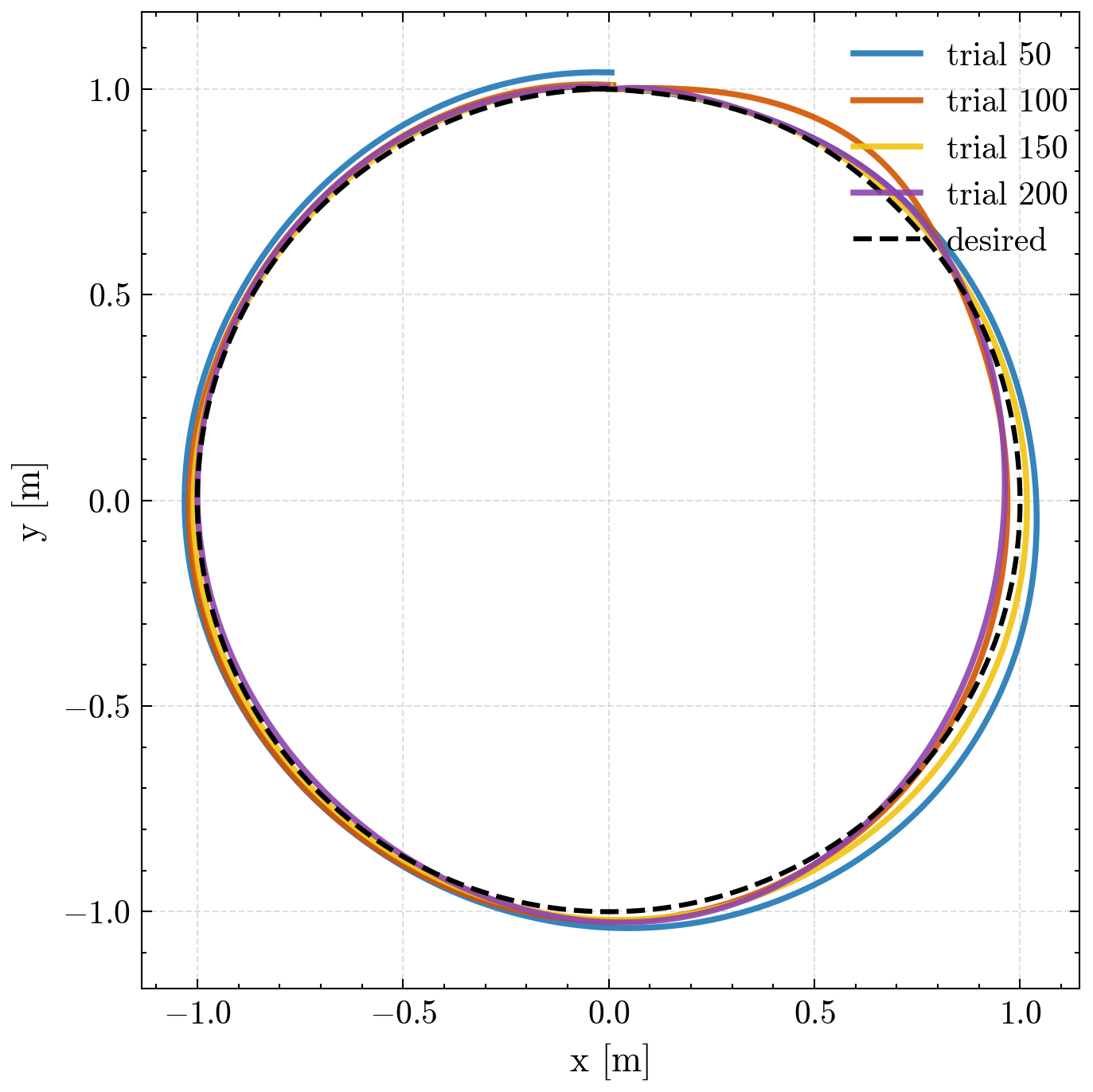}
        \label{fig:traj_c}
    }
    \caption{
    Trajectory evolution at representative training stages (Trials 50, 100, 150, and 200).
    (a) The \textit{DiffTune} baseline exhibits persistent steady-state error, indicating limited adaptability when optimization is restricted to controller gains.
    (b) \textit{Adjoint} substantially improves tracking performance, bringing the trajectory closer to the reference.
    (c) The proposed \textit{MetaTune} method achieves the tightest tracking accuracy, suggesting that network-based parameterization can effectively exploit joint gradient information for precise control.
    }
    \label{fig:traj_comparison}
\end{figure*}

\section{Experiments}

In this section, we evaluate the proposed MetaTune framework on a quadrotor platform to demonstrate joint auto-tuning of feedback controllers and disturbance observers. The quadrotor dynamics, control laws, and observer are fully differentiable, enabling end-to-end optimization via the proposed adjoint-based method.

\paragraph{Dynamics}

The quadrotor dynamics are described by
\begin{align*}
    \dot{p} &= v, &
    m\dot{v} &= m g e_{3} - f R e_{3} + d_{f}, \\
    \dot{R} &= R \Omega_{\times}, &
    J\dot{\Omega} &= -\Omega_{\times} J \Omega + \tau + d_{\tau},
\end{align*}
where $p, v \in \mathbb{R}^3$ and $R \in \mathrm{SO}(3)$ denote the position, velocity, and attitude of the quadrotor, and $\Omega \in \mathbb{R}^3$ is the body angular velocity. The mass and inertia matrix are denoted by $m$ and $J \in \mathbb{R}^{3 \times 3}$. The control inputs are the total thrust $f$ and body torque $\tau$, while $d_f$ and $d_\tau$ represent lumped aerodynamic disturbances. The operator $(\cdot)_\times$ denotes the skew-symmetric mapping.
Each task $\mathcal{T} \sim p(\mathcal{T})$ is defined by a wind disturbance profile and a minimum-snap reference trajectory\cite{martin2010true, mellinger2011minimum, richter2016polynomial}, with varying disturbance magnitudes, directions, and trajectories.

\paragraph{Geometric SE(3) Control with Disturbance Observer}
We employ a geometric tracking controller on $\mathrm{SE}(3)$~\cite{lee2010geometric} to generate the thrust $f$ and torque $\tau$, parameterized by axis-wise gains $k_p, k_v, k_R, k_\Omega \in \mathbb{R}^3$, collected as $\theta \in \mathbb{R}^{12}$. A nonlinear extended state observer (ESO)~\cite{sharma2021nonlinear, miao2024robust} 
estimates lumped disturbances in translation and rotation, with gains $\psi \in \mathbb{R}^6$. The full parameter vector is $\Theta = [\theta, \psi]^\top \in \mathbb{R}^{18}$.

MetaTune is implemented in Python with JAX~\cite{bradbury2018jax}, leveraging JIT compilation, vmap, and automatic differentiation to build a fully differentiable closed-loop simulator.

\subsection{Experimental Setup}

\subsubsection{\highlight[comment={R6404-C2/R6570-C4}]{Comparison Methods}}
We compare our adjoint-based approach against several forward-mode extensions derived from DiffTune\cite{cheng2024difftune}. The first baseline, \textit{DT-Base}, represents the original implementation which optimizes only fixed controller gains. To evaluate joint tuning, we extend forward sensitivity to both controller and observer parameters called \textit{DT-Fixed}. For time-varying settings, we implement two variants: \textit{DT-Adaptive (History)}, which propagates sensitivities once to accumulate gradients from past states, and \textit{DT-Adaptive (CTG)}, which performs a separate sensitivity computation at each update to capture the cost-to-go gradient. 

Our proposed methods, \textit{Adj-Fixed} and \textit{Adj-Adaptive}, leverage reverse-mode adjoint sensitivity to compute gradients. 
The distinction between forward sensitivity rollout and reverse-mode adjoint computation is summarized in Fig.~2 (Sec.~III). 
By integrating backward in time, these methods inherently capture the cost-to-go across the entire horizon.
\added[comment={R6385-C3}]{Both variants use the same backward adjoint sweep; the former aggregates node-wise gradients for a shared fixed gain, while the latter applies them directly to time-varying gains along the trajectory.}

\subsubsection{\highlight[comment={R6385-C2}]{Objective Function}}
We use a quadratic per-step objective defined over the system state, observer state, and control input,
\begin{equation}
\ell(x,\hat{x},u,r)
=
\|x - x_d(r)\|_{W_x}^2
+
\|\hat{x} - x\|_{W_{\hat{x}}}^2
+
\lambda_u \|u\|^2 ,
\end{equation}
The weighting matrices $W_x$ and $W_{\hat{x}}$ are diagonal and assign different penalties to state and observer components. For \textit{DiffTune}-based baselines, the observer-related entries of $W_{\hat{x}}$ are set to zero. The control penalty term $\lambda_u \|u\|^2$ serves as a regularization on control effort.

\subsubsection{\highlight[comment={R6404-C2}]{Learnable Policy Parameterization}}
The learnable policy is parameterized as a single-hidden-layer MLP with 128 ReLU units. Its input consists of the current observation, desired state, ESO state, the polynomial coefficients of the active trajectory segment, and the current time. The network outputs an 18-dimensional vector corresponding to the controller and observer gains. 
The resulting gains are updated at 20\,Hz and applied using zero-order hold between update instants.
To ensure physical feasibility, the raw network outputs are mapped through a sigmoid function and scaled to lie within predefined safety bounds. The policy is trained using the Adam optimizer with an initial learning rate of $10^{-3}$.

\subsection{Experimental Results in Differentiable Simulation}
\subsubsection{Physical Gradient Validation}

\begin{figure}[t]
    \centering
    \includegraphics[width=1.0\linewidth]{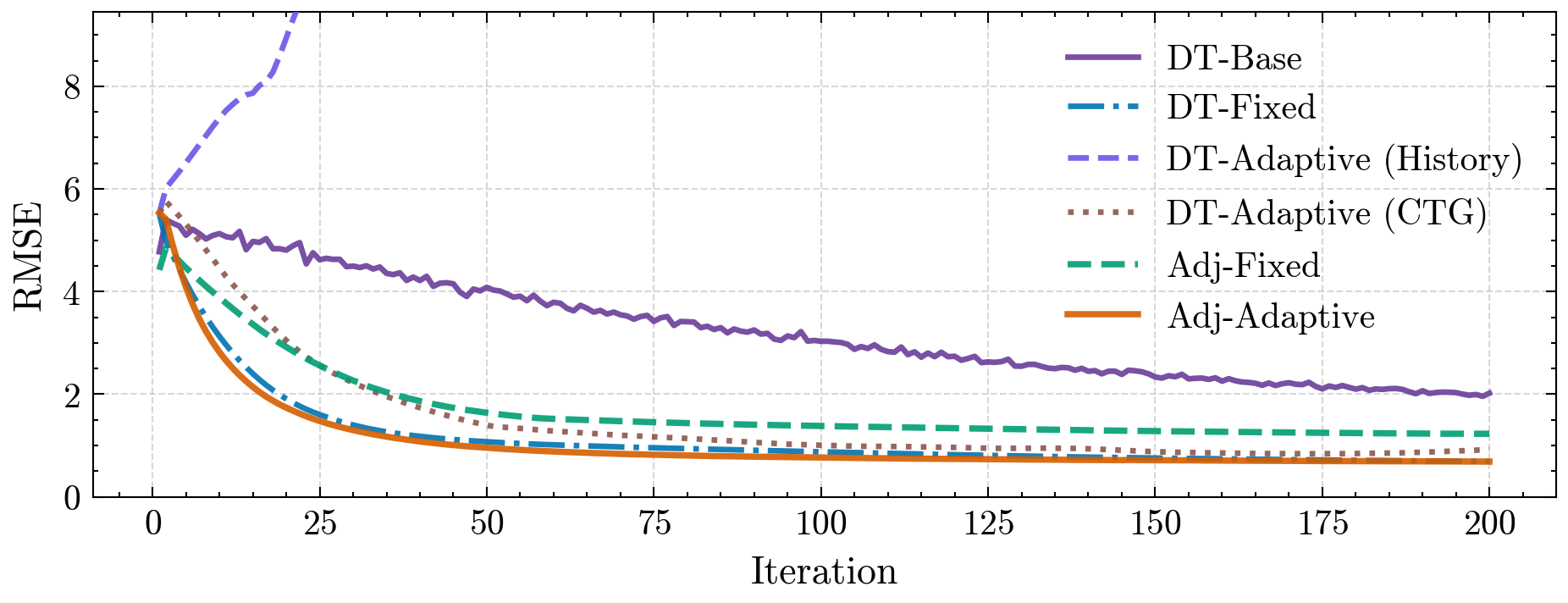}
    \caption{RMSE over training iterations.}
    \label{fig:loss_comparison}
\end{figure}

We evaluate the correctness and computational efficiency of different gradient computation strategies by directly optimizing controller and observer gains, and compare the proposed adjoint-based method with state-of-the-art differentiable tuning approaches. The training loss curves are shown in Fig.~\ref{fig:loss_comparison}, and the final RMSE together with gradient computation time are summarized in Table~\ref{tab:rmse_grad_time}.

\replaced[comment={R6385-C3}]{In the fixed-gain setting, \textit{Adj-Fixed} reduces runtime by more than 50\% relative to \textit{DT-Fixed}. 
More importantly, the adjoint runtime remains nearly unchanged when moving from fixed to adaptive gains, whereas the forward variants slow substantially. 
This efficiency gain benefits from the shared backward adjoint sweep, which avoids repeated gain-wise sensitivity rollouts.}
{While achieving comparable precision, the adjoint formulation significantly outperforms forward-mode differentiation in terms of efficiency. Even in the baseline case, Adj-Fixed reduces the computational footprint by over 50\% compared to DT-Fixed, highlighting its superior scaling laws when dealing with high-dimensional differential simulations.}

\replaced[comment={R6385-C3}]{The adaptive forward variants reveal the limitation of forward sensitivity rollout in the time-varying gain setting. 
\textit{DT-Adaptive (History)} suffers from unreliable temporal credit assignment, resulting in unstable training and a large RMSE. 
\textit{DT-Adaptive (CTG)} improves accuracy by explicitly recomputing cost-to-go gradients, but its runtime becomes prohibitively high. 
In contrast, \textit{Adj-Adaptive} achieves the lowest RMSE with a runtime close to \textit{Adj-Fixed}, demonstrating its effectiveness for coupled time-varying controller--observer tuning.}{A critical failure mode is observed in DT-Adaptive (History), where a single forward rollout fails to correctly handle temporal credit assignment, leading to training instabilities. 
Conversely, while DT-Adaptive (CTG) recovers performance by explicitly recomputing cost-to-go gradients, its prohibitive computational cost renders it impractical for real-time applications. 
In contrast, the Adj-Adaptive method achieves the lowest RMSE with orders-of-magnitude faster execution, proving its efficacy for coupled, time-varying controller--observer optimization.}

\input{method_compate_table}

\subsubsection{Policy-Augmented Adaptation and Generalization}
\label{sec:policy_learning}
We further evaluate the full MetaTune framework with the MLP-based gain scheduling policy trained in the differentiable JAX simulator using the adjoint-based meta-gradient described in Sec.~III. To assess the benefit of policy parameterization, we compare against gradient-based tuning methods that directly optimize gains without neural parameterization. The qualitative results in Fig.~\ref{fig:traj_comparison} support the following analysis.

The DiffTune baseline exhibits limited adaptability under disturbance, primarily due to its reliance on forward-mode sensitivity with fixed controller parameters. In this setting, parameter updates are driven by local instantaneous gradients and cannot effectively compensate for disturbance-induced model mismatch, particularly when observer dynamics remain constrained. As a result, disturbance effects accumulate over time and manifest as residual tracking error.

In contrast, adjoint-based tuning enables effective joint adaptation of controller and observer parameters by propagating cost-to-go sensitivities backward through the closed-loop system. This global temporal credit assignment allows disturbance-related errors to be correctly attributed to earlier parameter choices, leading to improved disturbance rejection and more consistent tracking behavior.

\subsection{Sim-to-Sim Transfer to PX4-Gazebo}

To evaluate sim-to-sim generalization, we deploy the policy trained in the differentiable JAX environment to a high-fidelity PX4-Gazebo hardware-in-the-loop (HIL) simulator without fine-tuning.
The HIL simulator includes more realistic aerodynamic effects, actuator dynamics, and low-level control constraints, and is used only for testing.
We compare MetaTune with an adjoint-optimized fixed-gain Baseline under different wind disturbances, flight speeds, and trajectory categories.

% \begin{figure}[t]
%     \centering
%     \subfloat[]{%
%         \includegraphics[width=0.48\columnwidth]{fig/disturbance.png}
%     }
%     \hfill
%     \subfloat[]{%
%         \includegraphics[width=0.48\columnwidth]{fig/disturbance_xyz_tracking.png}
%     }
%     \caption{Disturbance estimation and position tracking performance during hovering. 
%     (a) Estimated disturbance forces in the horizontal (XY) and vertical (Z) directions. 
%     (b) Position tracking along the X, Y, and Z axes.}
%     % TODO: This unreferenced figure may be replaced by an efficiency-scaling plot
%     % for the backward method versus horizon nodes and parameter dimension, or removed.
%     \label{fig:disturbance_tracking}
% \end{figure}

\subsubsection{Adaptation under High-Speed Flight}
\replaced[comment={R6404-C3}]{The results in Table~\ref{tab:results_mean_std} suggest that the benefit of MetaTune depends on the operating regime. At the low-speed setting of $2\,\text{m/s}$, the fixed-gain Baseline attains comparable, and in some weak-disturbance cases slightly lower, tracking errors than the adaptive policy. This can be attributed to the fact that the fixed gains are already well matched to the near-nominal flight condition, whereas online gain variation may introduce additional transient response when only limited adaptation is required.}
{The experimental results in Table~\ref{tab:results_mean_std} illustrate a clear trend: while MetaTune and the Baseline yield comparable performance at low speeds ($2\,\text{m/s}$), the adaptive policy becomes essential as flight regimes escalate in aggressiveness.}
As the speed and disturbance level increase, MetaTune demonstrates superior generalization, achieving an average RMSE reduction of approximately $15$--$20\%$ at higher velocities when compared to the fixed-gain Baseline.
The fixed-gain structure of the Baseline lacks the adaptability necessary to accommodate the nonlinear aerodynamic effects encountered during aggressive maneuvers. In contrast, MetaTune effectively compensates for model mismatches and transient aerodynamic effects in real time, demonstrating superior stability and responsiveness across diverse spatial trajectories.

\begin{figure*}[!b]
\centering

\subfloat[Circular trajectory]{%
    \includegraphics[width=0.32\textwidth]{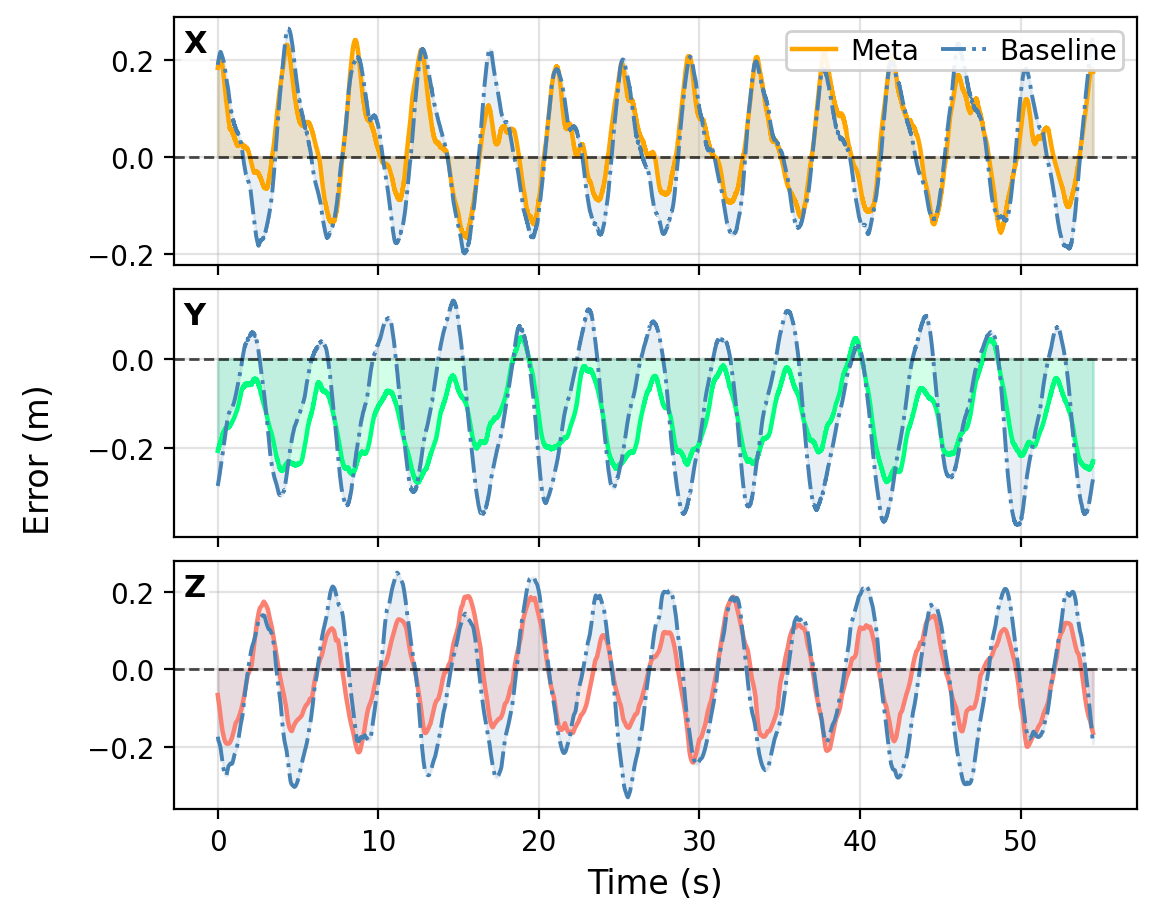}
    \label{fig:circle_error}
}
\hfill
\subfloat[Double-eight trajectory]{%
    \includegraphics[width=0.32\textwidth]{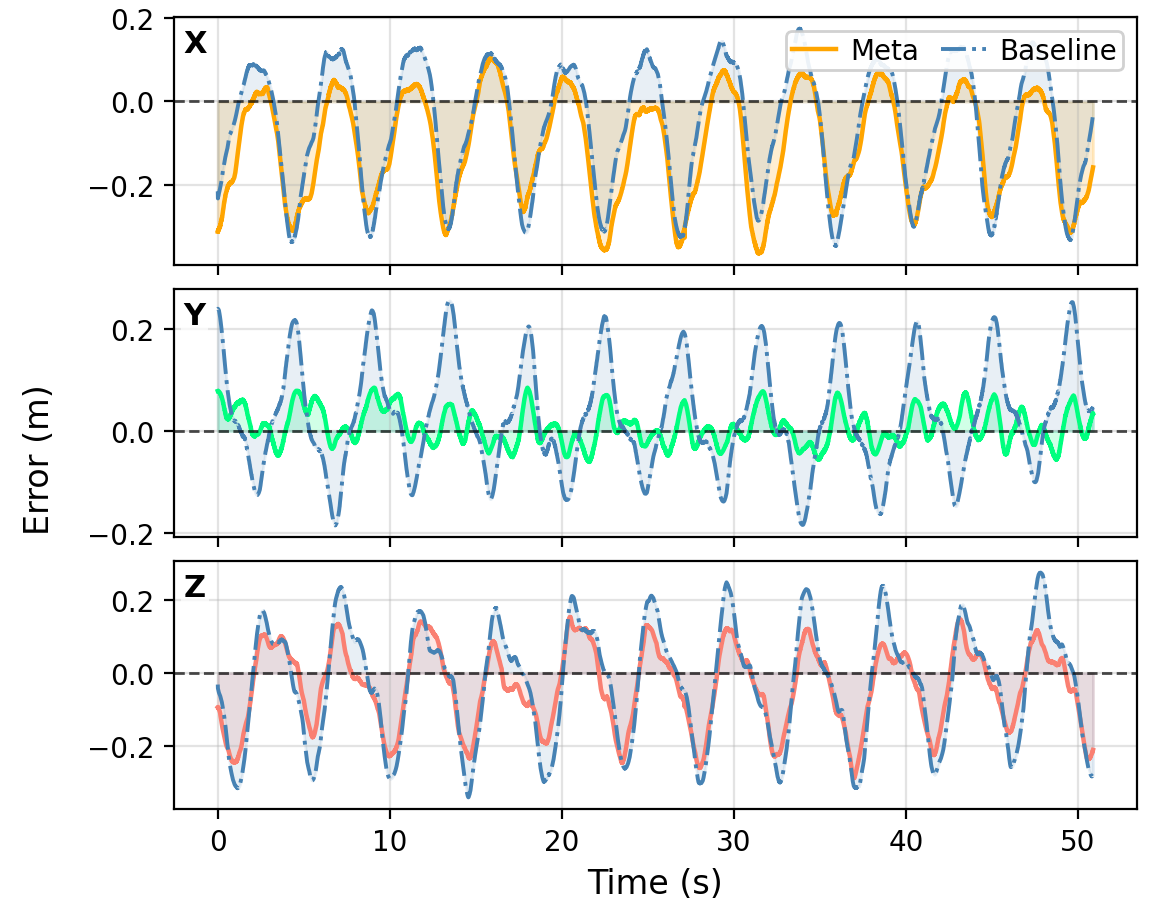}
    \label{fig:traj2_error}
}
\hfill
\subfloat[Planar figure-eight trajectory]{%
    \includegraphics[width=0.32\textwidth]{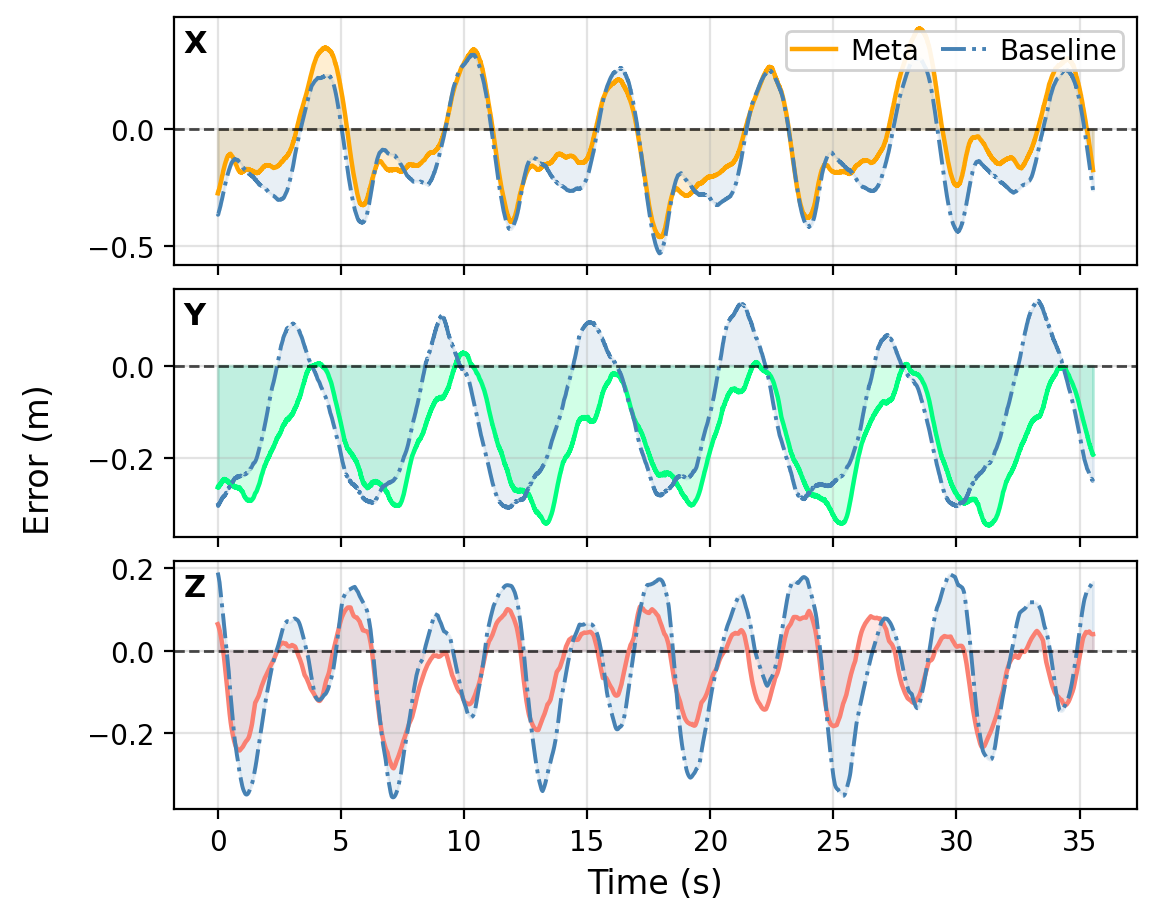}
    \label{fig:traj3_error}
}

\caption{Position error comparison under three trajectories.
The proposed Meta controller significantly reduces oscillation amplitude
compared with the baseline in X, Y, and Z axes.}
\label{fig:all_error}
\end{figure*}

\subsubsection{Robustness under Strong Disturbances}
The robustness advantage of MetaTune becomes more pronounced as external wind torque disturbances increase to 1\,N$\cdot$m and 2\,N$\cdot$m. While the Baseline controller exhibits significant performance degradation under these persistent loads—particularly during the dynamically demanding Figure-8 trajectory—the adaptive policy maintains a much more consistent tracking profile. In extreme scenarios combining high-velocity flight with maximum disturbance, MetaTune attains a reduction in RMSE of approximately $40\%$ when compared to the Baseline. This substantial improvement underscores the efficacy of the learned joint controller–observer adaptation. By tuning gains online in response to external torques, MetaTune effectively mitigates the tracking divergence commonly observed in fixed-parameter controllers when the disturbance magnitude exceeds the nominal design envelope.

\input{sim_result}

\subsubsection{Sim-to-Sim Generalization Capability}
The experimental results provide strong evidence of MetaTune's zero-shot transfer capability. Despite being trained exclusively in a simplified differentiable JAX environment, the policy maintains effective adaptation when deployed in the high-fidelity PX4-Gazebo simulator without any fine-tuning.

This robust performance across the sim-to-sim gap suggests that the physics-informed meta-learning framework captures fundamental closed-loop dynamics rather than overfitting to simulator-specific numerical traits. By internalizing the underlying physical relationships, MetaTune ensures that the learned gain-scheduling laws remain valid even when subjected to the unmodeled actuator dynamics and sensor noise inherent in more realistic simulation environments.

\section{Discussion and Conclusion}
In this work, we presented MetaTune, a differentiable framework for automated controller and observer joint tuning that enables efficient end-to-end optimization of a neural adaptive policy. Compared with forward sensitivity methods, MetaTune significantly reduces gradient computation overhead and improves scalability with respect to the horizon length and parameter dimension. Experiments in the high-fidelity PX4-Gazebo simulator show strong zero-shot generalization, achieving precision tracking across diverse trajectories and aerodynamic disturbances. Notably, MetaTune consistently outperforms fixed-gain baselines during high-speed flight and in windy conditions, highlighting the importance of state-dependent gain adaptation. These results suggest that grounding neural control policies in differentiable physics is a promising path toward robust robotic control.
\added[comment={R6385-C1}]{The present evaluation is limited to simulation and PX4-Gazebo HIL.}
\replaced[comment={R6385-C1}]{Future work includes real-flight tests and the incorporation of safety-critical constraints into the differentiable optimization framework to improve robustness and reliability in physical deployment.}{Future work will investigate broader real-world deployment and incorporate safety-critical constraints into the differentiable optimization framework to enhance robustness and reliability in safety-critical settings.}

\bibliographystyle{IEEEtran}
\bibliography{references}

\end{document}

%% file: algorithm_metatune.tex
\begin{algorithm}[t]
\caption{MetaTune Training Loop}
\label{alg:adjoint}
\begin{algorithmic}[1]
\State \textbf{Input:} Task dist. \( p(\mathcal{T}) \), initial params \( \xi \), learning rate \( \alpha \)
\State \textbf{Output:} Optimized parameters \( \xi^* \)
\While{not converged}
    \State Sample batch of tasks $\{\mathcal{T}^i\}_{i=1}^B \sim p(\mathcal{T})$
    \For{$t = 0$ \textbf{to} $T_{\text{episode}}$} \Comment{Forward Pass}
        \State Compute gains $\Theta_t$ via \eqref{eq:nn}
        \State Apply control $u_t$ \eqref{eq:controller} and update observer \eqref{eq:observer}
    \EndFor
    \For{$t = T_{\text{episode}}$ \textbf{to} $0$} \Comment{Backward Pass}
        \State Update adjoints $\lambda_{u,k}, \lambda_{\hat{x},k}, \lambda_{x,k}$ via \eqref{eq:lambda_u}--\eqref{eq:lambda_x}
        \State Accumulate gradients $\nabla_\xi L$ via chain rule \eqref{eq:chain}
    \EndFor
    \State Update $\xi$ using optimizer
\EndWhile
\end{algorithmic}
\end{algorithm}

%% file: method_compate_table.tex
\begin{table}[t]
\centering
\caption{Comparison of final RMSE and gradient computation time.}
\label{tab:rmse_grad_time}
\begin{tabular}{llcc}
\toprule
\textbf{Method} & \textbf{Variant} & \textbf{RMSE} & \textbf{Time [s]} \\
\midrule
\multirow{4}{*}{Forward} 
 & DT-Base & 1.56 & \textbf{$<10^{-3}$} \\
 & DT-Fixed & 0.69 & 0.57 \\
 & DT-Adaptive (History) & 15.58 & 1.52 \\
 & DT-Adaptive (CTG) & 0.92 & 17.81 \\
\midrule
\multirow{2}{*}{Backward} 
 & Adj-Fixed & 0.69 & 0.25 \\
 & Adj-Adaptive & \textbf{0.67} & 0.27 \\
\bottomrule
\end{tabular}
\end{table}

%% file: sim_result.tex
\begin{table*}[t]
\centering
\caption{Tracking performance under different velocities and wind conditions.}
\label{tab:results_mean_std}
\resizebox{\textwidth}{!}{%
\begin{tabular}{@{}llcccccc@{}}
\toprule
\multirow{2}{*}{\textbf{Vel.}} & \multirow{2}{*}{\textbf{Wind}}
& \multicolumn{2}{c}{\textbf{2D Circle}}
& \multicolumn{2}{c}{\textbf{3D Circle}}
& \multicolumn{2}{c}{\textbf{Figure-8}} \\
\cmidrule(lr){3-4} \cmidrule(lr){5-6} \cmidrule(l){7-8}
& & \textbf{Baseline} & \textbf{Ours} & \textbf{Baseline} & \textbf{Ours} & \textbf{Baseline} & \textbf{Ours} \\
\midrule
$2\,\mathrm{m/s}$ & $0\,\mathrm{N\cdot m}$
& $\mathbf{0.2397 \pm 0.0134}$ & $0.2415 \pm 0.0131$
& $\mathbf{0.2400 \pm 0.0120}$ & $0.2538 \pm 0.0160$
& $0.1819 \pm 0.0105$ & $\mathbf{0.1640 \pm 0.0139}$ \\
$2\,\mathrm{m/s}$ & $1\,\mathrm{N\cdot m}$
& $\mathbf{0.2323 \pm 0.0107}$ & $0.2369 \pm 0.0143$
& $\mathbf{0.2427 \pm 0.0103}$ & $0.2682 \pm 0.0160$
& $\mathbf{0.1454 \pm 0.0068}$ & $0.1659 \pm 0.0166$ \\
$2\,\mathrm{m/s}$ & $2\,\mathrm{N\cdot m}$
& $\mathbf{0.2290 \pm 0.0080}$ & $0.2505 \pm 0.0089$
& $\mathbf{0.2374 \pm 0.0148}$ & $0.2725 \pm 0.0104$
& $\mathbf{0.1481 \pm 0.0043}$ & $0.1684 \pm 0.0126$ \\
\midrule
$3\,\mathrm{m/s}$ & $0\,\mathrm{N\cdot m}$
& $0.2726 \pm 0.0147$ & $\mathbf{0.2314 \pm 0.0195}$
& $0.2172 \pm 0.0137$ & $\mathbf{0.1658 \pm 0.0092}$
& $0.3145 \pm 0.0142$ & $\mathbf{0.2824 \pm 0.0158}$ \\
$3\,\mathrm{m/s}$ & $1\,\mathrm{N\cdot m}$
& $0.2363 \pm 0.0144$ & $\mathbf{0.2311 \pm 0.0159}$
& $0.2149 \pm 0.0187$ & $\mathbf{0.1680 \pm 0.0072}$
& $0.3189 \pm 0.0093$ & $\mathbf{0.2793 \pm 0.0118}$ \\
$3\,\mathrm{m/s}$ & $2\,\mathrm{N\cdot m}$
& $0.2467 \pm 0.0131$ & $\mathbf{0.2199 \pm 0.0142}$
& $0.1985 \pm 0.0132$ & $\mathbf{0.1692 \pm 0.0087}$
& $0.3096 \pm 0.0127$ & $\mathbf{0.2774 \pm 0.0160}$ \\
\midrule
$4\,\mathrm{m/s}$ & $0\,\mathrm{N\cdot m}$
& $0.3465 \pm 0.0127$ & $\mathbf{0.2875 \pm 0.0126}$
& $0.3968 \pm 0.0185$ & $\mathbf{0.3683 \pm 0.0256}$
& $0.4252 \pm 0.0420$ & $\mathbf{0.3802 \pm 0.0226}$ \\
$4\,\mathrm{m/s}$ & $1\,\mathrm{N\cdot m}$
& $0.3299 \pm 0.0135$ & $\mathbf{0.2769 \pm 0.0124}$
& $0.3829 \pm 0.0108$ & $\mathbf{0.3249 \pm 0.0179}$
& $\text{crash}$ & $\mathbf{0.3881 \pm 0.0263}$ \\
$4\,\mathrm{m/s}$ & $2\,\mathrm{N\cdot m}$
& $0.3138 \pm 0.0176$ & $\mathbf{0.2778 \pm 0.0147}$
& $0.3454 \pm 0.0169$ & $\mathbf{0.3033 \pm 0.0178}$
& $\text{crash}$ & $\mathbf{0.4098 \pm 0.0149}$ \\
\bottomrule
\end{tabular}%
}
\end{table*}